\pdfoutput=1
\documentclass[11pt]{article}

\usepackage[preprint]{acl}

\usepackage{times}
\usepackage{latexsym}

\usepackage[T1]{fontenc}
\usepackage[utf8]{inputenc}

\usepackage{microtype}

\usepackage{inconsolata}

\usepackage[utf8]{inputenc}
\usepackage{booktabs}
\usepackage{graphicx}
\usepackage{CJK}
\usepackage{multirow}
\usepackage{color}
\usepackage{verbatim}
\usepackage{url}
\usepackage{enumitem}
\usepackage{amsmath}
\usepackage{diagbox}
\usepackage{flushend,cuted}
\usepackage{bbm}
\usepackage{xspace}
\usepackage{algorithm}
\usepackage{algpseudocode}
\usepackage{colortbl}
\usepackage{bbding}
\usepackage{amssymb}

\usepackage{listings}
\usepackage{ulem}
\usepackage{listings}
\usepackage{wrapfig}
\usepackage{soul}
\usepackage{makecell}

\usepackage{xcolor}

\newcommand{\kaiti}[1]{\begin{CJK*}{UTF8}{gkai} #1 \end{CJK*}}
\soulregister{\kaiti}7
\usepackage{pifont}
\usepackage{arydshln}
\usepackage{booktabs}

\title{Generalizing From Short to Long: Effective Data Synthesis for Long-Context Instruction Tuning}

\author{
    Wenhao Zhu$^{1}$\text{,} \textbf{Pinzhen Chen}$^{2}$\text{,} \textbf{Hanxu Hu}$^{4}$\text{,} Shujian Huang$^{1}$\text{,} \\
    \textbf{Fei Yuan}$^{3}$\text{,} \textbf{Jiajun Chen}$^{1}$\text{,} \textbf{Alexandra Birch}$^{2}$ \\
    $^{1}$ \text{National Key Laboratory for Novel Software Technology, Nanjing University} \\
    $^{2}$ \text{School of Informatics, University of Edinburgh } \\
    $^{3}$ \text{Shanghai Artificial Intelligence Laboratory} $^{4}$ \text{University of Zurich} \\
    \small\texttt{zhuwh@smail.nju.edu.cn}, \small\texttt{pinzhen.chen@ed.ac.uk}, \small\texttt{hanxu.hu@uzh.ch}, \small\texttt{huangsj@nju.edu.cn}, \\ \small\texttt{yuanfei@pjlab.org.cn}, \small\texttt{chenjj@nju.edu.cn}, \small\texttt{a.birch@ed.ac.uk} \\
}

\begin{document}
\maketitle

\begin{abstract}
Long-context modelling for large language models (LLMs) has been a key area of recent research because many real world use cases require reasoning over longer inputs such as documents.
The focus of research into modelling long context has been on how to model position and there has been little investigation into other important aspects of language modelling such as instruction tuning.
Long context training examples are challenging and expensive to create and use. In this paper, we investigate how to design instruction data for the post-training phase of a long context pre-trained model: how much and what type of context is needed for optimal and efficient post-training. 
Our controlled study reveals that models instruction-tuned on short contexts can effectively generalize to longer ones, while also identifying other critical factors such as instruction difficulty and context composition.
Based on these findings, we propose \textit{context synthesis}, a novel data synthesis framework that leverages off-the-shelf LLMs to generate extended background contexts for high-quality instruction-answer pairs.
Experiment results on the document-level benchmark (\textsc{LongBench}) demonstrate that our proposed approach outperforms previous instruction synthesis approaches and comes close to the performance of human-annotated long-context instruction data\footnote{The project will be available at: \url{https://github.com/NJUNLP/context-synthesis}.}.
\end{abstract}

\section{Introduction}
Recent advances in large language models (LLMs) have significantly extended the length of the context that they are able to ingest by addressing the problems of efficient attention and encoding of positions~\cite{su2024roformer,peng2024yarn,fu2024data,dubey2024llama}.
However, for better performance in downstream long-context tasks such as document-level question answering and summarization~\cite{shaham2023zeroscrolls,bai2024longbench,karpinska2024one}, these models still require instruction-tuning.
Although previous work has looked at synthetic instructions, we still do not understand how to best leverage existing instructions for optimal use with a long-context model.
This is a critical challenge, as it is difficult to create high quality synthetic long-context instructions.

\begin{figure}
    \centering
    \includegraphics[width=0.8\linewidth]{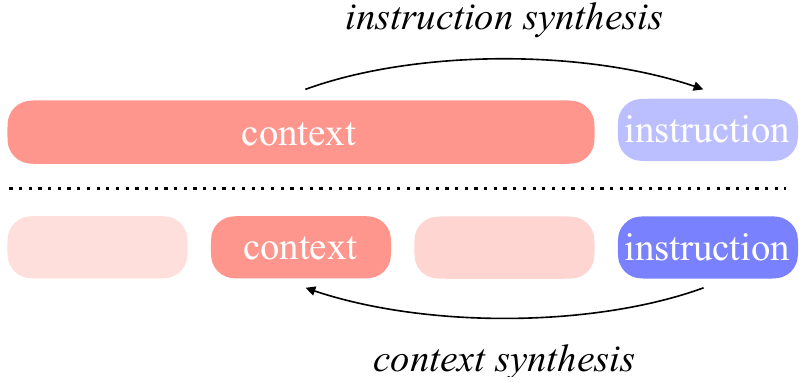}
    \caption{Illustration of two long-context instruction data synthesis frameworks: \textit{instruction synthesis} and \textit{context synthesis} (ours). The light-colored blocks indicate potentially lower-quality components in the synthesized data samples.}
    \label{fig:illustration}
\end{figure}

Initial work on long-context instructions leverages off-the-shelf LLMs to generate instruction-answer pairs from existing long text passages~\cite{bai2024longalign, dubey2024llama}.
While this method prioritizes context length but overlooks other critical aspects of the synthetic data such as quality, difficulty and diversity.
These aspects are inherently constrained by an underlying paradox: the synthesized data quality relies on an LLM that can understand the lengthy input text---which is the problem it is tackling in the first place.

This paper first presents a pilot study on artificial needle-in-a-haystack tests~\cite{kamradt2023needle, hsieh2024ruler} which allows for rigorous control of different aspects of the instruction data.
This study yields three key findings: (1) instruction quality plays a crucial role in model performance; (2) models instruction-tuned on short contexts can generalize to much longer ones; (3) training with evidence embedded in distracting content helps models develop robust information extraction abilities.

We leverage these findings to design a novel instruction data synthesis approach called ``\textit{context synthesis}'' (Figure~\ref{fig:illustration}) and test it on naturally occurring tasks. 
Specifically, we prompt off-the-shelf LLMs to generate background context from existing instruction-answer pairs.
This approach offers three advantages: (1) in contrast to previous work which synthesizes instructions and target outputs, our synthetic data only forms part of the input to the model rather like back-translation for machine translation~\cite{sennrich2016improving}, preserving the quality of instructions and outputs.
(2) by generating background contexts, we can seamlessly integrate both supporting evidence and distracting information into a coherent narrative.
(3) our approach enables control over context through expansion and concatenation to harness the benefits of training on longer sequences.

We conduct experiments on real-world tasks from \textsc{LongBench}~\cite{bai2024longbench} with two base models \texttt{LLaMA2-7B-64K}~\cite{bai2024longalign} and \texttt{LLaMA3.1-8B-128K}~\cite{dubey2024llama}.
Experimental results demonstrate that our context synthesis approach significantly outperforms the instruction synthesis methods and comes close to the performance of fine-tuning with oracle human-annotated long-context instruction data.
Further analysis comparing instruction tuning with and without context reveals that the performance gains from previous instruction synthesis methods depend minimally on the paired long context, indicating their limitations in ensuring instruction-context alignment.
In contrast, training instructions with our synthesized input context enables LLMs to learn effective patterns for context utilization.
Furthermore, our instruction-tuned models demonstrate robust generalization to unseen tasks from other document-level benchmarks including \textsc{RULER}~\cite{hsieh2024ruler} and \textsc{ZeroScrolls}~\cite{shaham2023zeroscrolls}.
To summarize, our contributions are as follows: 
\begin{itemize}[itemsep=0pt, topsep=0pt, parsep=0pt]
\item We identify key factors in data synthesis for long-context instruction tuning through a controlled study, including instruction quality, context composition, and context length.
\item We propose a novel data synthesis method, context synthesis, that addresses the key factors by generating tailored background context for high-quality instructions.
\item Experimental results demonstrate that our approach outperforms the previous instruction synthesis approach and achieves performance close to using human-annotated data.
\item We devise an analytical tool which reveals the limitations of existing synthesis approaches in data quality.
\end{itemize}

\section{Related Work}
\paragraph{Towards Long-context LLMs}
To enable LLMs to support longer context, most previous research focuses on the pre-training stage by modifying rotary position embeddings (RoPE)~\cite{su2024roformer, peng2024yarn} along with continued pre-training on longer sequences~\cite{chen2023extending,rozière2024codellama,chen2024longlora,peng2024yarn,xiong2024effective,fu2024data}.
However, while previous studies observe that current LLMs often struggle with long-context instruction following~\cite{shaham2023zeroscrolls}, few studies systematically investigate the instruction-tuning stage for long-context tasks.
In this paper, we aim to understand critical factors in data synthesis for long-context instruction tuning.

\noindent\paragraph{Long-context Instruction Data Synthesis}
A key challenge for long-context tasks is the scarcity of long-context instruction data.
Initially, \citet{chen2024longlora} released LongAlpaca, a publicly available long-context instruction dataset, though its annotation process is unclear.
Recent works propose ``instruction synthesis'' to prompt LLMs to generate instruction-answer pairs from long documents~\cite{bai2024longalign,xiong2024effective,dubey2024llama}, exemplified by the open-source dataset LongAlign.
However, this approach raises concerns about instruction quality, task complexity~\cite{chen2024essential} and task diversity~\cite{quan2024language}, as current LLMs themselves still struggle with understanding long-context messages.
Subsequently, \citet{chen2024essential} developed a multi-agent workflow to improve synthesis quality for multi-hop reasoning instruction data (LongMIT).
Meanwhile, a contrasting perspective from \citet{gao2024train} suggests that synthetic long-context data offers minimal benefits, arguing that using standard-length general instruction data, e.g., ShareGPT~\cite{chiang2023vicuna} suffices.
Our work, however, reveals limitations in existing instruction synthesis approaches and confirms the importance of long-context instruction data through a novel and effective synthesis framework.

\begin{figure*}[ht!]
    \centering
    \includegraphics[width=0.95\textwidth]{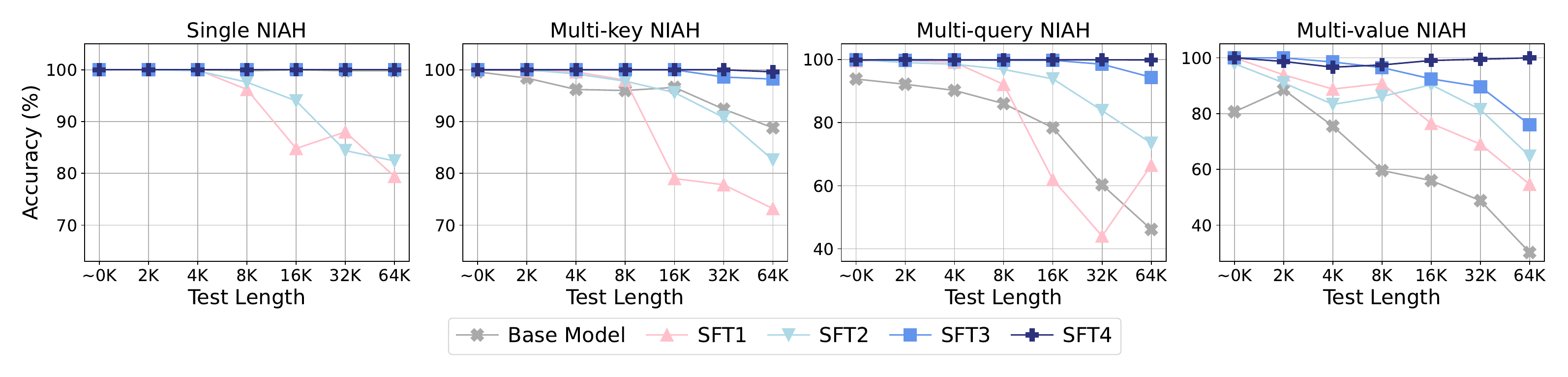}
    \caption{Impact of varying instruction tuning configurations on long-context performance. The detailed differences between these configurations is presented in Table~\ref{tab:explanation}. Test length ``$\sim$0k''  means test contexts containing only the relevant information (needle) without any additional content.}
   \label{fig:niah}
\end{figure*}

\noindent\paragraph{Long-context Tasks Evaluation}
Early research focuses on measuring perplexity on long-context data~\cite{chen2023extending,peng2024yarn} and evaluating passkey retrieval~\cite{fu2024data}.
Recent developments introduce real-world benchmarks such as \textsc{ZeroSCROLLS}, \textsc{LongBench}, which assess practical capabilities through document-level tasks.
In this work, we establish our hypotheses via a controllable passkey retrieval task and then evaluate our data synthesis approach on real-world benchmarks to demonstrate practical effectiveness.

\section{Pilot Study}
\label{sec:pilot_study}

\subsection{Concept Definitions}
We first introduce essential concepts used throughout the discussion.
An instruction data instance consists of two components: a \textbf{prompt} and an \textbf{answer}.
For general-purpose instruction data, the prompt contains only an instruction that requires the LLM to generate an answer based on its parametric knowledge.
For context-aware instruction data, the prompt contains both a \textbf{context passage} and an \textbf{instruction}.
Typically, the instruction requests a part of the context (i.e., the relevant or evidence context), while the remaining irrelevant context may distract the model from generating an accurate response.
In the following analysis, we systematically control the instruction data to understand the key factors that influence the effectiveness of instruction-tuned LLMs on long-context tasks.

\subsection{Experiment Design}

\paragraph{Testing Scenario}
We adopt the needle-in-a-haystack\footnote{The target ``needles'' embedded in the context are randomly generated words and UUIDs (Universally Unique Identifiers).} (NIAH) test~\cite{kamradt2023needle} for our analysis, which challenges LLMs to locate evidence buried within long contexts.
\begingroup
\renewcommand{\arraystretch}{1.1}
\begin{table}[ht]
\centering
\resizebox{0.48\textwidth}{!}{
\footnotesize
\begin{tabular}{ccccc}
\toprule
\multirow{2}{*}{\textbf{Model}} & \multirow{2}{*}{\makecell{\textbf{General} \\ \textbf{Instruction Data}}} & \multirow{2}{*}{\makecell{\textbf{Context-aware} \\ \textbf{Instruction Data}}} & \multicolumn{2}{c}{\textbf{Context Composition}} \\
 & &  & \textbf{rel. (needle)} & \textbf{irr. (haystack)} \\
\midrule
\texttt{Base}  &  -           & -                  & -                      & -                  \\
\texttt{SFT1}  & \checkmark             & -                  & -                      & -                  \\
\texttt{SFT2}  & \checkmark             & \checkmark         & \texttt{uuid sent}     & -                  \\
\texttt{SFT3}  & \checkmark             & \checkmark         & \texttt{uuid sent}     & \texttt{1k essay}  \\
\texttt{SFT4}  & \checkmark             & \checkmark         & \texttt{uuid sent}     & \texttt{64k essay} \\
\bottomrule
\end{tabular}}
\caption{Overview of instruction-tuning configurations for experiments in Figure~\ref{fig:niah}. The context in context-aware instruction data consists of two parts: relevant context (a sentence containing the target UUID, usually called "needle") and irrelevant context (an essay serving as distractive information, usually called "haystack"). We present the data format for these tasks in Appendix~\ref{sec:niah-data-format}.}
\label{tab:explanation}
\end{table}
\endgroup
We also incorporate its extended variants proposed in \textsc{RULER} benchmark~\cite{hsieh2024ruler}: multi-key, multi-query and multi-value NIAH tests, which requires LLMs to identify multiple pieces of information amid distracting contexts.
We pick these tasks for our pilot study due to their controllable context composition and length. % , allowing for variable control.
We conduct experiments using \texttt{LLaMA2-7B-64k}\footnote{\texttt{LLaMA2-7B-64k} is an extended version of \texttt{LLaMA2}~\cite{touvron2023llama2}, continued pre-trained by Bai et al. to support longer contexts with a context window extended to 64k tokens.} as the base model for instruction-tuning.

\noindent\paragraph{Instruction Data Control}
Table~\ref{tab:explanation} summarizes our experimental configurations for instruction-tuning.
For \texttt{SFT1}, we only use general instruction data from ShareGPT~\cite{vicuna2023}.
The remaining \texttt{SFT} models incorporate specialized context-aware instruction data with varying data composition and length.
We evaluate three training conditions: without distracting context (\texttt{SFT2}), with short distracting context (\texttt{SFT3}), and with long distracting context (\texttt{SFT4}).
To create the context-aware instruction data, we construct 200 NIAH-test-style training samples\footnote{In these training samples, we use the Paul Graham Essays~\cite{kamradt2023needle} as the haystack, consistent with the testing scenario, but incorporating newly generated contents as needles.} for each of the four subtasks, resulting in a total of 800 samples.

\subsection{Empirical Results and Insights}
\paragraph{The Need for Context-aware Instruction Data}
In Figure~\ref{fig:niah}, we compare the base model with the model trained with only general instruction data (\texttt{SFT1}) and with additional specialized context-aware instruction data (\texttt{SFT2}, \texttt{SFT3}, \texttt{SFT4}).
Results show that as the test length increases, \texttt{SFT1} exhibit significant performance divergence and even underperforms the \texttt{Base Model} on single, multi-key and multi-query NIAH tasks.
In contrast, incorporating a pinch of targeted context-aware instruction data (\texttt{SFT2}, \texttt{SFT3}, \texttt{SFT4}) leads to substantial performance improvements. 
This observation leads to our first insight: \textit{unlocking pre-trained long-context LLMs' potential requires specialized instruction data beyond general instruction data.}

\noindent\paragraph{Short-Context Training Generalizes to Long Contexts}
Next, we compare models trained with context-aware instruction data of varying context compositions and data lengths (\texttt{SFT2}, \texttt{SFT3}, \texttt{SFT4}).
As shown in Figure~\ref{fig:niah}, training with distracting context (\texttt{SFT3}, \texttt{SFT4}) significantly improves the model's performance with longer contexts.
In contrast, models trained solely with relevant context (\texttt{SFT2}) may develop shortcuts, leading to performance degradation when exposed to distracting information. 
Notably, \texttt{SFT3} maintains performance above 90\% across all NIAH tasks, even when test lengths reach 32k - far beyond its instruction-tuning context length, suggesting a potentially more efficient and cost-effective training approach.
This observation leads to our second insight: \textit{training with both evidence and distracting contexts, even short ones, is crucial for developing robust generalization to longer contexts}.

\noindent\paragraph{Training with Long Contexts Remains Optimal}
Although \texttt{SFT3} approaches the performance of \texttt{SFT4} with much shorter contexts, \texttt{SFT4} demonstrate almost perfect performance across all evaluated tasks.
The performance gap between the two models is particularly pronounced at the maximum test length in the most challenging task (Multi-value NIAH).
This highlights the important role of context length in instruction data for achieving optimal performance.
This observation leads to our third insight: \textit{training with long-context instruction data achieves optimal performance, especially for more challenging long-context tasks}.

\section{Applying Insights to Real-World Tasks}
In real-world scenarios, manually annotating long-context instruction data is both complex and labor-intensive, making the synthesis of context-aware instruction data a critical research challenge.
Based on insights from our pilot study, we propose a novel method called \textit{context synthesis} (\S\ref{sec:context}) and discuss the limitations of existing instruction synthesis approach (\S\ref{sec:instruction}).
Additionally, we propose an analytic tool for measuring the quality of synthesized data, particularly the coherence between contexts and instructions (\S\ref{sec:context-free-tuning}).

\subsection{Previous Approach: Instruction Synthesis}
\label{sec:instruction}
The existing approach to data synthesis, known as ``instruction synthesis'' (Figure~\ref{fig:illustration}), starts with long text passages and uses off-the-shelf LLMs to generate instruction-answer pairs based on the given text~\cite{bai2024longalign}.
This method focuses primarily on context length when constructing instruction data for long-context tasks, while overlooking other critical factors such as instruction quality. 
The automatically generated instructions often lack quality guarantees, and the off-the-shelf LLMs may not have sufficient capacity to effectively process long contexts, compromising the coherence between contexts and synthetic instructions.
Furthermore, the source passages may lack complex or contradictory information that could serve as distractors, limiting the model's robustness in handling noisy real-world scenarios.

\subsection{Proposed Approach: Context Synthesis}
\label{sec:context}
Unlike the previous approach, our method starts from existing instruction-answer pairs and synthesizes the corresponding context.
This approach makes the synthetic content merely part of the input to the model, 
thus prioritizing the quality of the instruction-answer pairs because they are naturally occurring. 
Additionally, this design enables control over the context: we can deliberately incorporate complex distractors while maintaining tight coupling between instructions and contexts.
Furthermore, by generating manageable moderate-length contexts, our synthesis process avoids the paradox of relying on a strong long-context LLM for instruction data synthesis.

\begin{figure}[htb]
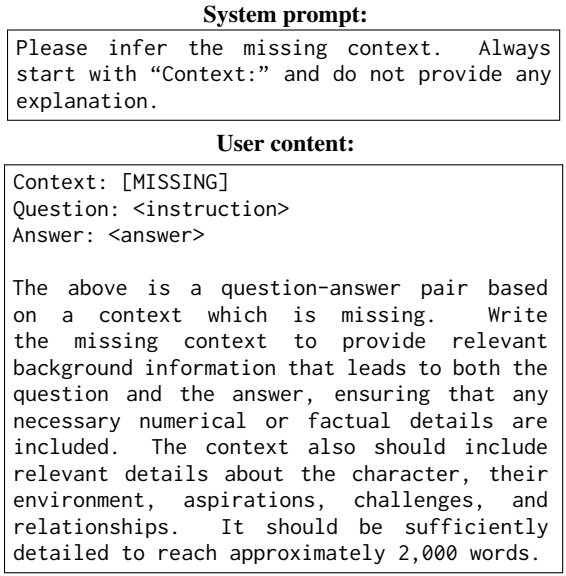

  \centering\small
  \textbf{System prompt:}
  \vspace{1ex}
  \noindent\framebox{%
  \parbox{0.44\textwidth}{
  \texttt{Please infer the missing context. Always start with ``Context:'' and do not provide any explanation.}
  }%
  }
  \vspace{1ex}
  \textbf{User content:}
  \vspace{1ex}
  \noindent\framebox{%
  \parbox{0.44\textwidth}{
  \texttt{Context: [MISSING]\\Question: <instruction>\\Answer: <answer>\\
  \\
  The above is a question-answer pair based on a context which is missing. Write the missing context to provide relevant background information that leads to both the question and the answer, ensuring that any necessary numerical or factual details are included. The context also should include relevant details about the character, their environment, aspirations, challenges, and relationships. It should  be sufficiently detailed to reach approximately 2,000 words.}
  }
  }
\caption{Our prompt template for synthesizing context from instruction-answer pairs. The template takes an instruction-answer pair as input, where \texttt{<instruction>} and \texttt{<answer>} are replaced with the actual instruction and answer text. The system prompt ensures the output follows a consistent format, while the user content guides the LLM to generate context that supports the given instruction-answer pair.}
\label{fig:prompt-template-context-synthesis}
\end{figure}

\noindent\paragraph{Instruction Collection}
At first, we collect instruction-answer pairs for context synthesis.
In this paper, our collection considers two key requirements.
The instructions should require in-context knowledge to answer, preventing models from relying solely on their parametric knowledge\footnote{We initially experiment with instruction-answer pairs from Alpaca~\cite{wang2023instruct}, but find them ineffective as those instructions primarily test parametric knowledge.}.
Additionally, we aim to employ a setup that allows for controlled comparison between synthetic and human-annotated data.
To meet these requirements, we source our instruction-answer pairs from human-annotated context-aware datasets while synthesizing new contexts rather than using the original paired ones.

\noindent\paragraph{Synthesizing Evidence Context}
Starting with a collection of instruction-answer pairs, we prompt off-the-shelf LLMs to synthesize background context for them.
The prompt template we designed is shown in Figure~\ref{fig:prompt-template-context-synthesis}.
Our prompt guides the LLM engine to generate context that supports the instruction and ensures the answer can be derived from the context.
From our observations\footnote{We present some example cases in Appendix \ref{sec:example-synthesized-context}.}, the synthesized context maintains a strong and coherent relationship with the given instruction while naturally including detailed and distracting information.

\noindent\paragraph{Extending Synthesized Context}
While training on shorter-context data generalize to longer ones, our pilot study still indicates that training on longer contexts yields better performance.
To extend context length, we investigate two approaches, depending on whether the extended portion remains coherent with the evidence context.
One approach is increasing the word count requirements in the prompt.
We set the target context length to 2,000 words, as recent studies have shown that current LLMs struggle to generate content beyond this length~\cite{bai2024longwriter,pham2024suri,quan2024language}.

The second approach is incorporating incoherent text into the context, similar to the method described in ~\citet{dubey2024llama}.
Specifically, we take contexts synthesized for other instruction-answer pairs and integrate them into the current one.
Our experiments demonstrate that both approaches are effective, with their combination yielding the best results.

\subsection{Quality Measurement for Synthetic Data}
\label{sec:context-free-tuning}
Beyond proposing a framework for data synthesis, we also introduce an analytical tool to verify the quality of synthesized data, especially the coherence between instruction and context.
For ideal context-aware instruction data, there should be strong interdependence between the context and instruction.
Based on this principle, we propose using the performance gap between context-free and context-included tuning as a quality indicator.
If context-free tuning yields results similar to context-included tuning, it indicates potential quality issues in the synthesized data.
We apply this analytical tool to reveal limitations in existing instruction synthesis approaches in our experiments (Section~\ref{sec:analysis}).

\section{Experiments}
\label{sec:main_results}
\subsection{Experimental Setting}

\paragraph{Evaluation Benchmark}
We conduct evaluations on \textsc{LongBench}\footnote{\url{https://github.com/THUDM/LongBench}}, focusing on three representative long-context tasks: single-document question-answering, multiple-documents question-answering and summarization\footnote{Despite being widely used in long-context evaluation, it has been argued that summarization tasks may suffer from position bias, as models can utilize the few leading sentences to achieve decent performance~\cite{nallapati2017summarunner,li2024long}.} (Table~\ref{tab:longbench}).
\begin{table}[ht]
\scriptsize
\renewcommand{\arraystretch}{1.2}
\centering
\resizebox{0.45\textwidth}{!}{
\begin{tabular}{clcc}
\toprule
\textbf{Type}                          & \hspace{0.5cm}\textbf{Dataset}  &\textbf{Size}  & \textbf{Metric}  \\
\hline
\rowcolor{gray!15}
\multicolumn{4}{c}{\textit{Artificial Tasks}} \\
% \hline
\multirow{4}{*}{Passkey Retrieval}     & Single NIAH      & 500 & EM      \\
                                       & Multi-key NIAH   & 500 & EM      \\
                                       & Multi-query NIAH & 500 & EM      \\
                                       & Multi-value NIAH & 500 & EM      \\
\hline
\rowcolor{gray!15}
\multicolumn{4}{c}{\textit{Real-world Tasks}} \\
% \hline
\multirow{2}{*}{Single-doc QA}         & NarrativeQA    & 200  & F1      \\
                                       & Qasper         & 200  & F1      \\
% \hline
\cline{2-4} 
\multirow{3}{*}{Multi-doc QA}         & HotpotQA        & 200  & F1      \\
                                      & 2WikiMultihopQA & 200  & F1      \\
                                      & Musique         & 200  & F1      \\
\cline{2-4}
\multirow{3}{*}{Summarization}        & GovReport       & 200  & Rouge-L \\
                                      & QMSum           & 200  & Rouge-L \\
                                      & MultiNews       & 200  & Rouge-L \\
\bottomrule
\end{tabular}
}
\caption{Long-context evaluation benchmark used in our experiments. In the pilot study (Section~\ref{sec:pilot_study}), we use tasks from \textsc{RULER}. In main experiments (Section~\ref{sec:main_results}), we use real-world tasks from \textsc{LongBench}. The context length distribution of these tasks is shown in Figure~\ref{fig:length_and_score}.}
\label{tab:longbench}
\end{table}
With the open-sourced codebase~\cite{bai2024longbench}, we conduct zero-shot evaluations using greedy decoding with the provided templates and metrics.

\noindent\paragraph{Base Models}
We use \texttt{LLaMA2-7B-64k}\footnote{\url{https://huggingface.co/THUDM/LongAlign-7B-64k-base}}
and \texttt{LLaMA3.1-8B-128k}\footnote{\url{https://huggingface.co/meta-llama/Llama-3.1-8B}} as base models for instruction-tuning.
Both models are pre-trained to support extended context windows (64k and 128k tokens respectively).

\noindent\paragraph{Instruction Data}
For specialized context-aware instruction data, we randomly sample 200 instructions from each subtask's training set (totalling 1.6k instructions) for context synthesis\footnote{We set the number of concatenated contexts to ten and report performance with context concatenation unless otherwise stated. Performance with different concatenation sizes is reported in Appendix~\ref{sec:number-of-concatenated-context}.}.
To ensure a fair comparison, we use the corresponding contexts from these selected samples for instruction synthesis (identical sample size and domain).
We employ \texttt{GPT4o-mini}\footnote{\texttt{GPT4o-mini} refers to \texttt{gpt-4o-mini-2024-07-18}, used with default temperature and top-p settings.}~\cite{openai2024gpt4ocard} as the data synthesis engine\footnote{In the analysis section, we report perform of using other models as the synthesis engine, such as \texttt{LongWriter-8B}~\cite{bai2024longwriter} and \texttt{Qwen2.5-72B-Instruct}~\cite{qwen2025qwen25}.}.
We also compare our approach with open-source long-context instruction data as strong baselines, including LongAlpaca, LongAlign and LongMIT. 
For general-purpose instruction data, we employ ShareGPT for \texttt{LLaMA2} and UltraChat for \texttt{LLaMA3.1}\footnote{The results of using ShareGPT with \texttt{LLaMA3.1} are reported in Appendix~\ref{sec:llama3.1-sharegpt}.}.

\noindent\paragraph{Instruction-tuning Configuration}
We adopt \textit{LongAlign}\footnote{\url{https://github.com/THUDM/LongAlign}} as the codebase for instruction-tuning.
Training is performed on a single node with 8$\times$H800 GPUs.
Detailed training configuration is reported in Appendix~\ref{sec:instruction-tuning-details}.

\begingroup
\renewcommand{\arraystretch}{1.2} % Default value: 1
\begin{table*}[ht]
\centering
\scriptsize
\resizebox{\textwidth}{!}{
\begin{tabular}{p{4cm}cccccccccc}
\hline
\textbf{\hspace{1cm}Instruction Data}     & \#\textbf{Size} & \textbf{NarrativeQA} & \textbf{Qasper}      & \textbf{HotpotQA}    & \textbf{2WikiQA}    & \textbf{MuSiQue}     & \textbf{GovReport} & \textbf{QMSum}       & \textbf{MultiNews} & \textbf{Avg.}        \\
\hline
\texttt{\hspace{1cm}LLaMA2-7B}        &       &                     &                &                &                &                 &                &                &                &                      \\
ShareGPT~\cite{vicuna2023}                & 89.3k & 15.88               & 23.18          & 22.15          & 19.34          & 9.90            & 29.57          & 24.19          & 26.44          & 21.33                \\
\hdashline     
\textit{open-source long-context data}    &       &                     &                &                &                &                 &                &                &                &                      \\
+ LongAlpaca~\cite{chen2024longlora}      & 12.0k & 18.00               & 26.82          & 24.37          & 20.05          & 11.00           & 30.00          & 25.75          & 26.85          & 22.86                \\
+ LongAlign~\cite{bai2024longalign}       & 9.9k  & 17.90               & 33.21          & 30.63          & 23.91          & 11.70           & 29.81          & 22.77          & 26.61          & 24.57                \\
+ LongMIT~\cite{chen2024essential}        & 64.4k & 19.74               & 34.10          & 41.27          & 26.03          & 21.90           & 28.12          & 23.94          & 26.26          & 27.67                     \\
\hdashline 
\textit{controlled comparison}            &            &                &                &                &                &                 &                &                &                &                      \\
+ Instruction Synthesis~\cite{bai2024longalign} & 1.6k & 19.33          & 31.74          & 28.26          & 24.27          & 14.74           & 29.05          & 23.85          & 26.02          & 24.66                \\
+ Context Synthesis (ours)                & 1.6k       & \textbf{27.10} & \textbf{40.55} & \textbf{47.92} & \textbf{32.69} & \textbf{29.09}  & \textbf{32.03} & \textbf{27.17} & \textbf{30.81} & \textbf{33.42}       \\
\hline
\texttt{\hspace{1cm}LLaMA3.1-8B}     &            &                &                &                &                 &                &                &                &                &                      \\
UltraChat~\cite{ding2023enhancing}        & 515.3k     & 22.45          & 28.12          & 24.00          & 19.38           & 9.08           & 30.24          & 26.18          & 27.36          & 23.35                \\
\hdashline
\textit{open-source long-context data}    &            &                &                &                &                 &                &                &                &                &                      \\
+ LongAlpaca~\cite{chen2024longlora}      & 12.0k      & 22.80          & 28.88          & 22.94          & 22.33           & 10.57          & 30.64          & 26.18          & 27.55          & 23.99                \\
+ LongAlign~\cite{bai2024longalign}       & 9.9k       & 17.02          & 25.70          & 11.54          & 12.22           & 7.82           & 29.65          & 25.65          & 27.23          & 19.60                \\
+ LongMIT~\cite{chen2024essential}        & 64.4k      & 25.08          & 37.71          & 34.35          & 29.68           & 18.78          & 29.83          & 25.47          & 27.42          & 28.54                \\
\hdashline
\textit{controlled comparison}            &            &                &                &                &                 &                &                &                &                &                      \\
+ Instruction Synthesis~\cite{bai2024longalign} & 1.6k & 24.39          & 29.32          & 30.26          & 21.68           & 14.99          & 29.85          & 25.60          & 27.02          & 25.39                \\
+ Context Synthesis (ours)                & 1.6k       & \textbf{32.74} & \textbf{45.30} & \textbf{59.73} & \textbf{44.28}  & \textbf{32.20} & \textbf{35.82} & \textbf{27.79} & \textbf{30.70} & \textbf{38.57}       \\
\hline
\end{tabular}
}
\caption{This table illustrates model performance between using general instruction data alone and using additional long-context instruction data (rows with `+'), and compares our context synthesis approach against the previous instruction synthesis approach in a controlled setting. We also report performance using other open-source long-context instruction data for reference. Bold text denotes the highest score among instruction-tuned models.}
\label{tab:compare}
\end{table*}
\endgroup

\begin{figure*}[ht]
    \centering
    \includegraphics[width=0.95\linewidth]{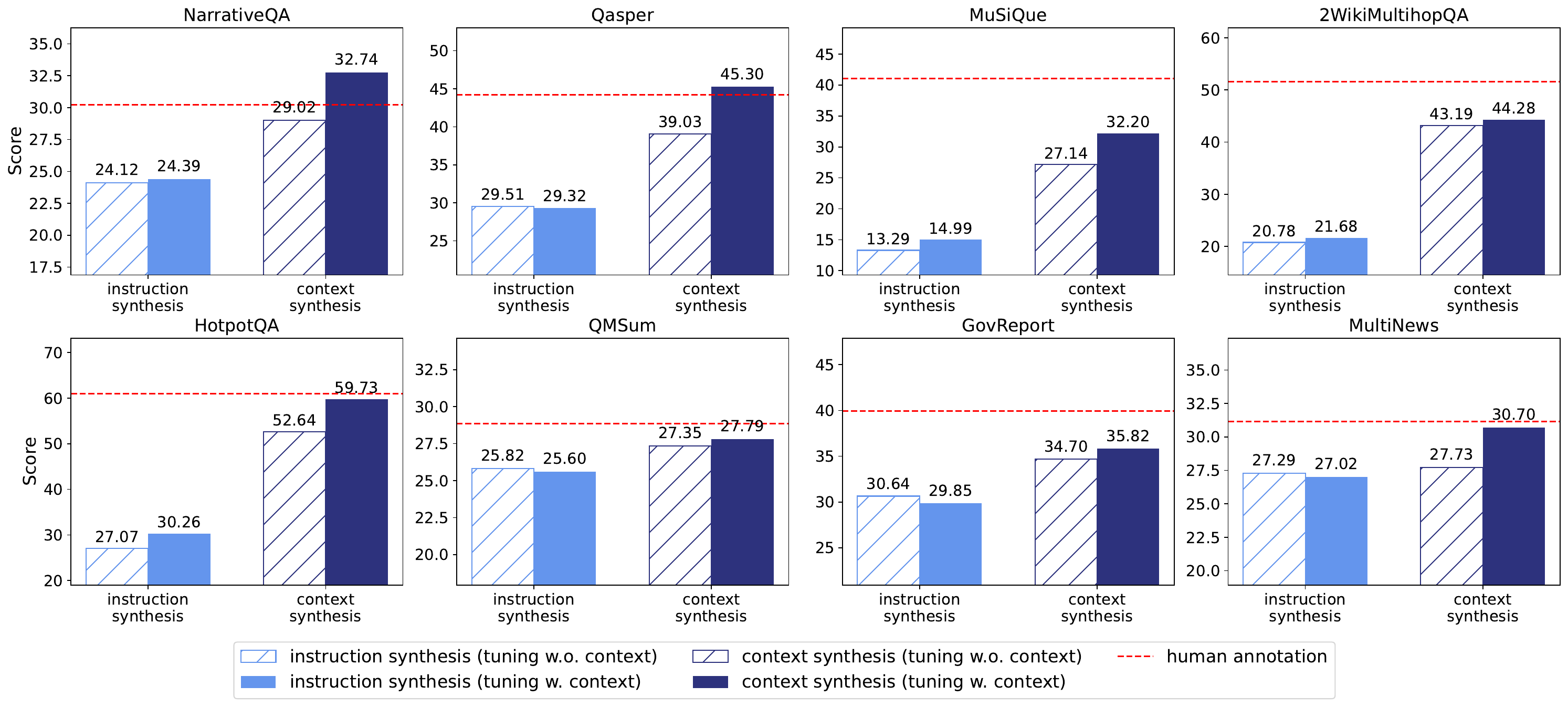}
    \caption{In this figure we compare model performance after instruction-tuning contrasting instruction synthesis with our approach of context synthesis. In both cases, we compare tuning without context (diagonal lines) with tuning with context (solid bars). We also illustrate the gap between synthesized data and oracle human-annotated data (red dotted line). Experiments are conducted with \texttt{LLaMA3.1-8B}.}
    \label{fig:human_gap}
\end{figure*}
\subsection{Main Results}
\label{sec:main}

\paragraph{Context Synthesis Approach Significantly Improves Long-context Performance}
In Table~\ref{tab:compare}, we first demonstrate the benefits of using long-context instruction data over using general-purpose instruction data alone.
We then compare our context synthesis approach against the instruction synthesis approach, while also reporting performance using open-source long-context instruction data for reference, though their construction processes are either transparent or do not allow for a fair comparison.
Results show that incorporating our synthesized instruction data significantly improves the model's long-context performance, achieving the highest scores across all tasks.

\noindent\paragraph{Context Synthesis Outperforms Instruction Synthesis on All Evaluated Tasks}
While instruction synthesis, which prioritizes long text lengths, shows improvement over using general instruction data alone in most cases, it yields suboptimal performance compared to the context synthesis approach (Table~\ref{tab:compare}).
Our experiments with LongAlpaca and LongAlign on the cutting-edge LLM (\texttt{LLaMA3.1}) show almost no performance improvement, which aligns with findings in \citet{gao2024train}.
We attribute this performance gap to the instruction quality, rather than suggesting context-aware instruction data is unnecessary, which will be quantitatively analyzed later (Section~\ref{sec:analysis}).

\noindent\paragraph{Context Synthesis Is Closing the Gap with Human-annotated Data}
The primary objective for synthesizing long-context instruction data is to achieve the effectiveness of human-annotated data with minimal cost.
Our experiments also enable a controlled comparison between data synthesis strategies and human-annotated long-context instruction data (Figure~\ref{fig:human_gap}).
Results show that the instruction synthesis approach significantly underperforms human-annotated data.
In contrast, our context synthesis approach achieves comparable or superior performance to human-annotated training data on single-document question-answering and summarization tasks.
Nevertheless, a performance gap persists in multi-document question-answering tasks, suggesting the need for further research in this direction.

\section{Analysis}
\label{sec:analysis}

\paragraph{Revealing Limitations of Instruction Synthesis with Our Analytical Tool}
As introduced in Section~\ref{sec:context-free-tuning}, we employ context-free tuning as an analytical tool to assess the quality of synthesized instruction data (Figure~\ref{fig:human_gap}).
Notably, augmenting synthesized instructions with long text messages yields no additional improvements, suggesting that additional context information, although long, fails to provide meaningful learning signals.
In contrast, pairing instructions with our synthesized context leads to performance gains across different tasks, highlighting the compatibility between instructions and corresponding synthesized contexts.

\begin{figure*}[ht]
    \centering
    \includegraphics[width=0.9\linewidth]{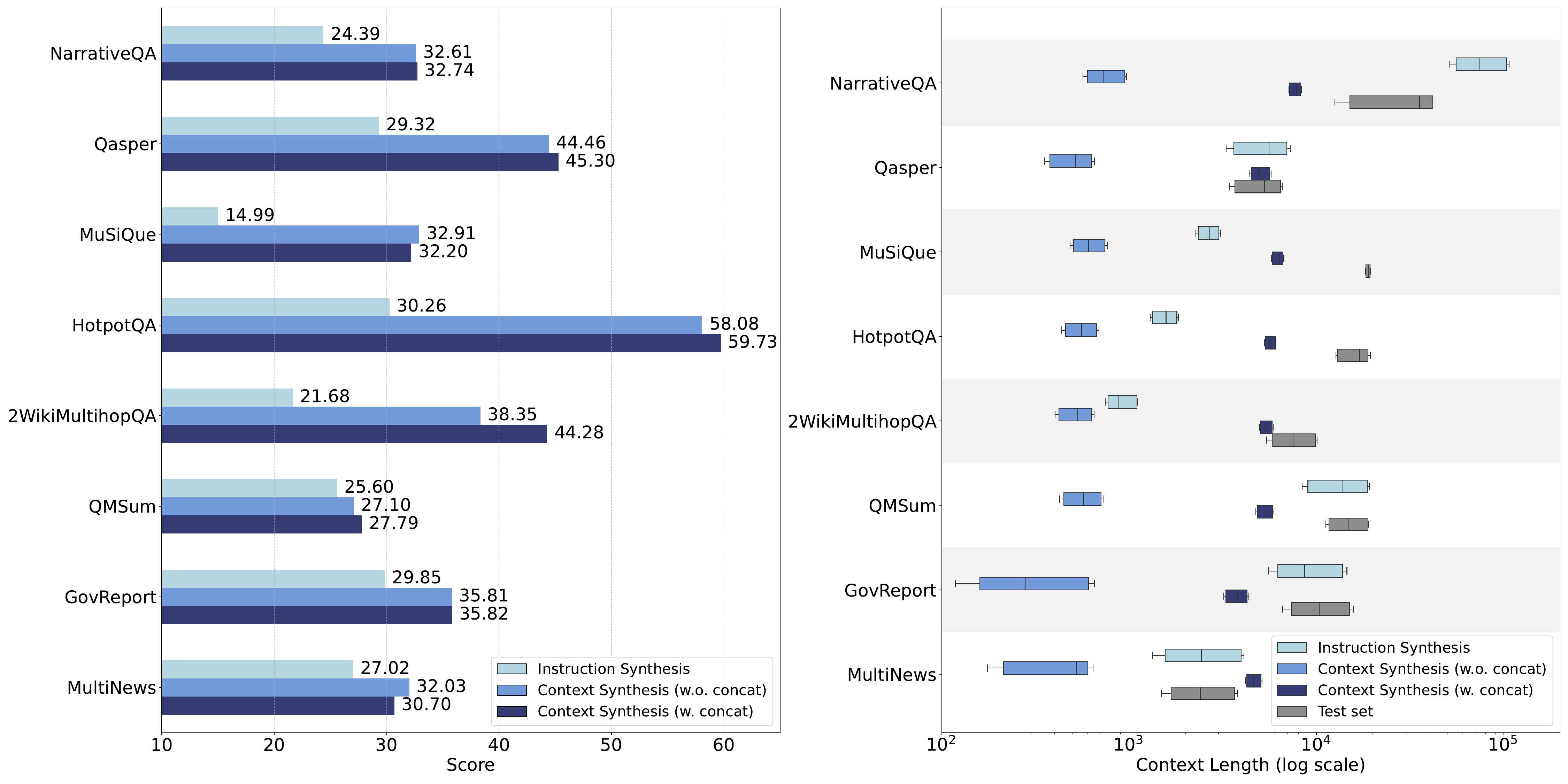}
    \caption{In the left panel, we present a task-wise performance comparison of different synthesis strategies. In the right panel, we display the context length distribution of different synthesis strategies against test sets across different tasks.}
    \label{fig:length_and_score}
\end{figure*}

\noindent\paragraph{Analysis of Length Generalization During Instruction Tuning}
Figure~\ref{fig:length_and_score} presents a comparative analysis of downstream task performance and context length distribution.
\begingroup
\renewcommand{\arraystretch}{1.3} % Default value: 1
\begin{table}[ht]
\footnotesize
\centering
\resizebox{0.48\textwidth}{!}{
\begin{tabular}{cccc}
\toprule
\textbf{Synthesis Engine} & \textbf{Single-Doc QA} & \textbf{Multi-Doc QA} & \textbf{Summ.} \\
\midrule
\texttt{LongWriter-8B} & 38.85 & 44.68 & 30.79 \\
\texttt{Qwen2.5-72B}   & 38.30 & 44.90 & 31.53 \\
\hdashline
\texttt{GPT4o-mini}    & 39.02 & 45.40 & 31.44 \\
\bottomrule
\end{tabular}
}
\caption{Average performance of using different LLMs as context synthesis engines across three groups of downstream tasks. All experiments are conducted with \texttt{LLaMA3.1-8B}.}
\label{tab:coherent}
\end{table}
\endgroup
Even without context concatenation, our model trained on shorter contexts generalizes effectively to long-context tasks.
These findings validate our observations on length generalization.
This also reveals an efficient instruction-tuning recipe that achieves strong performance with reduced data length.

\noindent\paragraph{Comparative Analysis of Different LLMs as Data Engines}
We evaluate various LLMs as data engines for context synthesis, including both proprietary models like GPT4o-mini and open-source models such as \texttt{LongWriter-8B} and \texttt{Qwen2.5-72B-instruct}.
Experimental results in Table~\ref{tab:coherent} demonstrate comparable performance across different engines, validating the robustness of our context synthesis approach.

\noindent\paragraph{Evaluating Generalization Capability on Unseen Tasks}
Moreover, we check the generalization capability of our data synthesis approach.
We evaluate our instruction-tuned model on tasks that were not seen during training, including the most challenging variant of NIAH task (multi-value NIAH), and \textsc{ZeroSCROLLS}\footnote{For \textsc{ZeroSCROLLS}, we follow \citet{dubey2024llama} and report numbers on the validation set. For QuALITY we report exact match, for SQuALITY we report ROUGE-L.}.
As shown in Table~\ref{tab:unseen}, our data synthesis approach demonstrates strong generalization to these unseen tasks, consistently outperforming the baseline that uses only general instruction data.

\begingroup
\renewcommand{\arraystretch}{1.3}
\begin{table}[]
\footnotesize
\centering
\resizebox{0.5\textwidth}{!}{
\begin{tabular}{lccc}
\toprule
\multirow{2}{*}{\textbf{\hspace{0.25cm}Instruction Data}} & \textbf{NIAH}       & \multicolumn{2}{c}{\textbf{ZeroSCROLLS}}   \\
                                        & \textbf{Multi-Value} & \textbf{QuALITY}  & \textbf{SQuALITY}                  \\
\midrule
UltraChat                 & 91.57                  & 71.43                     & 23.02                    \\
+ our synthesis data      & \textbf{96.54} (+4.97) & \textbf{76.19} (+4.76)    & \textbf{23.65} (+0.63)   \\
\bottomrule
\end{tabular}
}
\caption{Results on unseen tasks from \textsc{RULER} and \textsc{ZeroSCROLLS}. NIAH results show the average across different tested lengths. Numbers in parentheses indicate performance improvements. All experiments are conducted with \texttt{LLaMA3.1-8B}.}
\label{tab:unseen}
\end{table}
\endgroup

\section{Conclusion}
This work investigates effective data synthesis for long-context instruction-tuning.
Through a pilot study on controllable needle-in-a-haystack tasks, we identify that instruction difficulty, context composition, and context length all play crucial roles.
Based on these insights, we propose a novel synthesis approach called ``context synthesis''.
Experiment results on document-level question-answering and document-level summarization tasks demonstrate that our method not only outperforms the previous instruction synthesis approach but also achieves comparable performance to oracle human-annotated data.
Furthermore, our approach shows robust generalization to unseen tasks not covered during data synthesis. 
Additionally, we quantitatively assess instruction-context coherence, revealing new insights for designing effective long-context instruction data.

\section*{Limitations}
Our work has certain limitations.
While we evaluate on document-level question-answering and summarization tasks, these may not fully cover all real-world scenarios.
We plan to extend our scope as more practical long-context benchmarks emerge.
Additionally, as our context synthesis approaches require context-aware instructions as a starting point, an automated verification framework would be beneficial to filter such instructions from large instruction pools - a direction we leave for future work.
Moreover, while experiments in this study focuses on LLMs with quadratic attention mechanisms, we notice the recent trend of developing large language models with linear attention mechanisms. 
This shift may raise new research questions about long-context modeling, which we plan to explore in future work.

\section*{Acknowledgement}
We would like to thank Qipeng Guo, Yijun Yang for their suggestions and feedback.
This work is supported by National Science Foundation of China (No. 62376116, 62176120), the Fundamental Research Funds for the Central Universities (No. 2024300507) and research project of Nanjing University-China Mobile Joint Institute.
This project has also received funding from UK Research and Innovation (UKRI) under the UK government's Horizon Europe funding guarantee (grant numbers 10039436 and 10052546).
Wenhao Zhu is also supported by China Scholarship Council (No.202306190172).
Shujian Huang is the corresponding author. 

\normalem
\bibliography{custom}

\begin{thebibliography}{30}
\providecommand{\natexlab}[1]{#1}

\bibitem[{Bai et~al.(2024{\natexlab{a}})Bai, Lv, Zhang, He, Qi, Hou, Tang, Dong, and Li}]{bai2024longalign}
Yushi Bai, Xin Lv, Jiajie Zhang, Yuze He, Ji~Qi, Lei Hou, Jie Tang, Yuxiao Dong, and Juanzi Li. 2024{\natexlab{a}}.
\newblock \href {https://aclanthology.org/2024.findings-emnlp.74} {{L}ong{A}lign: A recipe for long context alignment of large language models}.
\newblock In \emph{Findings of the Association for Computational Linguistics: EMNLP 2024}.

\bibitem[{Bai et~al.(2024{\natexlab{b}})Bai, Lv, Zhang, Lyu, Tang, Huang, Du, Liu, Zeng, Hou, Dong, Tang, and Li}]{bai2024longbench}
Yushi Bai, Xin Lv, Jiajie Zhang, Hongchang Lyu, Jiankai Tang, Zhidian Huang, Zhengxiao Du, Xiao Liu, Aohan Zeng, Lei Hou, Yuxiao Dong, Jie Tang, and Juanzi Li. 2024{\natexlab{b}}.
\newblock \href {https://aclanthology.org/2024.acl-long.172} {{L}ong{B}ench: A bilingual, multitask benchmark for long context understanding}.
\newblock In \emph{Proceedings of the 62nd Annual Meeting of the Association for Computational Linguistics (Volume 1: Long Papers)}.

\bibitem[{Bai et~al.(2024{\natexlab{c}})Bai, Zhang, Lv, Zheng, Zhu, Hou, Dong, Tang, and Li}]{bai2024longwriter}
Yushi Bai, Jiajie Zhang, Xin Lv, Linzhi Zheng, Siqi Zhu, Lei Hou, Yuxiao Dong, Jie Tang, and Juanzi Li. 2024{\natexlab{c}}.
\newblock \href {https://arxiv.org/abs/2408.07055} {Longwriter: Unleashing 10,000+ word generation from long context llms}.

\bibitem[{Chen et~al.(2023)Chen, Wong, Chen, and Tian}]{chen2023extending}
Shouyuan Chen, Sherman Wong, Liangjian Chen, and Yuandong Tian. 2023.
\newblock \href {https://arxiv.org/abs/2306.15595} {Extending context window of large language models via positional interpolation}.

\bibitem[{Chen et~al.(2024{\natexlab{a}})Chen, Qian, Tang, Lai, Liu, Han, and Jia}]{chen2024longlora}
Yukang Chen, Shengju Qian, Haotian Tang, Xin Lai, Zhijian Liu, Song Han, and Jiaya Jia. 2024{\natexlab{a}}.
\newblock \href {https://openreview.net/forum?id=6PmJoRfdaK} {Longlo{RA}: Efficient fine-tuning of long-context large language models}.
\newblock In \emph{The Twelfth International Conference on Learning Representations}.

\bibitem[{Chen et~al.(2024{\natexlab{b}})Chen, Chen, Qin, Guo, Lv, Zou, Che, Yan, Chen, and Lin}]{chen2024essential}
Zhi Chen, Qiguang Chen, Libo Qin, Qipeng Guo, Haijun Lv, Yicheng Zou, Wanxiang Che, Hang Yan, Kai Chen, and Dahua Lin. 2024{\natexlab{b}}.
\newblock \href {https://arxiv.org/abs/2409.01893} {What are the essential factors in crafting effective long context multi-hop instruction datasets? insights and best practices}.

\bibitem[{Chiang et~al.(2023{\natexlab{a}})Chiang, Li, Lin, Sheng, Wu, Zhang, Zheng, Zhuang, Zhuang, Gonzalez, Stoica, and Xing}]{chiang2023vicuna}
Wei-Lin Chiang, Zhuohan Li, Zi~Lin, Ying Sheng, Zhanghao Wu, Hao Zhang, Lianmin Zheng, Siyuan Zhuang, Yonghao Zhuang, Joseph~E. Gonzalez, Ion Stoica, and Eric~P. Xing. 2023{\natexlab{a}}.
\newblock \href {https://lmsys.org/blog/2023-03-30-vicuna/} {Vicuna: An open-source chatbot impressing gpt-4 with 90\%* chatgpt quality}.

\bibitem[{Chiang et~al.(2023{\natexlab{b}})Chiang, Li, Lin, Sheng, Wu, Zhang, Zheng, Zhuang, Zhuang, Gonzalez, Stoica, and Xing}]{vicuna2023}
Wei-Lin Chiang, Zhuohan Li, Zi~Lin, Ying Sheng, Zhanghao Wu, Hao Zhang, Lianmin Zheng, Siyuan Zhuang, Yonghao Zhuang, Joseph~E. Gonzalez, Ion Stoica, and Eric~P. Xing. 2023{\natexlab{b}}.
\newblock \href {https://lmsys.org/blog/2023-03-30-vicuna/} {Vicuna: An open-source chatbot impressing gpt-4 with 90\%* chatgpt quality}.

\bibitem[{Ding et~al.(2023)Ding, Chen, Xu, Qin, Hu, Liu, Sun, and Zhou}]{ding2023enhancing}
Ning Ding, Yulin Chen, Bokai Xu, Yujia Qin, Shengding Hu, Zhiyuan Liu, Maosong Sun, and Bowen Zhou. 2023.
\newblock \href {https://aclanthology.org/2023.emnlp-main.183} {Enhancing chat language models by scaling high-quality instructional conversations}.
\newblock In \emph{Proceedings of the 2023 Conference on Empirical Methods in Natural Language Processing}.

\bibitem[{Dubey et~al.(2024)Dubey, Jauhri, Pandey, Kadian, Al-Dahle, Letman, Mathur, Schelten, Yang, Fan et~al.}]{dubey2024llama}
Abhimanyu Dubey, Abhinav Jauhri, Abhinav Pandey, Abhishek Kadian, Ahmad Al-Dahle, Aiesha Letman, Akhil Mathur, Alan Schelten, Amy Yang, Angela Fan, et~al. 2024.
\newblock \href {https://arxiv.org/pdf/2407.21783} {The llama 3 herd of models}.
\newblock \emph{arXiv preprint arXiv:2407.21783}.

\bibitem[{Fu et~al.(2024)Fu, Panda, Niu, Yue, Hajishirzi, Kim, and Peng}]{fu2024data}
Yao Fu, Rameswar Panda, Xinyao Niu, Xiang Yue, Hannaneh Hajishirzi, Yoon Kim, and Hao Peng. 2024.
\newblock \href {https://arxiv.org/abs/2402.10171} {Data engineering for scaling language models to 128k context}.

\bibitem[{Gao et~al.(2024)Gao, Wettig, Yen, and Chen}]{gao2024train}
Tianyu Gao, Alexander Wettig, Howard Yen, and Danqi Chen. 2024.
\newblock \href {https://arxiv.org/abs/2410.02660} {How to train long-context language models (effectively)}.

\bibitem[{Hsieh et~al.(2024)Hsieh, Sun, Kriman, Acharya, Rekesh, Jia, and Ginsburg}]{hsieh2024ruler}
Cheng-Ping Hsieh, Simeng Sun, Samuel Kriman, Shantanu Acharya, Dima Rekesh, Fei Jia, and Boris Ginsburg. 2024.
\newblock \href {https://openreview.net/forum?id=kIoBbc76Sy} {{RULER}: What{\textquoteright}s the real context size of your long-context language models?}
\newblock In \emph{First Conference on Language Modeling}.

\bibitem[{Kamradt(2023)}]{kamradt2023needle}
Gregory Kamradt. 2023.
\newblock \href {https://github.com/gkamradt/LLMTestNeedleInAHaystack/tree/main} {Needle in a haystack - pressure testing llms}.

\bibitem[{Karpinska et~al.(2024)Karpinska, Thai, Lo, Goyal, and Iyyer}]{karpinska2024one}
Marzena Karpinska, Katherine Thai, Kyle Lo, Tanya Goyal, and Mohit Iyyer. 2024.
\newblock \href {https://arxiv.org/abs/2406.16264} {One thousand and one pairs: A "novel" challenge for long-context language models}.

\bibitem[{Li et~al.(2024)Li, Zhang, Do, Yue, and Chen}]{li2024long}
Tianle Li, Ge~Zhang, Quy~Duc Do, Xiang Yue, and Wenhu Chen. 2024.
\newblock \href {https://arxiv.org/abs/2404.02060} {Long-context llms struggle with long in-context learning}.
\newblock \emph{Preprint}, arXiv:2404.02060.

\bibitem[{Nallapati et~al.(2017)Nallapati, Zhai, and Zhou}]{nallapati2017summarunner}
Ramesh Nallapati, Feifei Zhai, and Bowen Zhou. 2017.
\newblock Summarunner: a recurrent neural network based sequence model for extractive summarization of documents.
\newblock In \emph{Proceedings of the Thirty-First AAAI Conference on Artificial Intelligence}.

\bibitem[{OpenAI(2024)}]{openai2024gpt4ocard}
OpenAI. 2024.
\newblock \href {https://arxiv.org/abs/2410.21276} {{GPT-4o} system card}.

\bibitem[{Peng et~al.(2024)Peng, Quesnelle, Fan, and Shippole}]{peng2024yarn}
Bowen Peng, Jeffrey Quesnelle, Honglu Fan, and Enrico Shippole. 2024.
\newblock \href {https://openreview.net/forum?id=wHBfxhZu1u} {Ya{RN}: Efficient context window extension of large language models}.
\newblock In \emph{The Twelfth International Conference on Learning Representations}.

\bibitem[{Pham et~al.(2024)Pham, Sun, and Iyyer}]{pham2024suri}
Chau~Minh Pham, Simeng Sun, and Mohit Iyyer. 2024.
\newblock \href {https://aclanthology.org/2024.findings-emnlp.94/} {{S}uri: Multi-constraint instruction following in long-form text generation}.
\newblock In \emph{Findings of the Association for Computational Linguistics: EMNLP 2024}.

\bibitem[{Quan et~al.(2024)Quan, Tang, Yu, Yang, Liu, Gao, Tu, Zhang, Zhou, and Lin}]{quan2024language}
Shanghaoran Quan, Tianyi Tang, Bowen Yu, An~Yang, Dayiheng Liu, Bofei Gao, Jianhong Tu, Yichang Zhang, Jingren Zhou, and Junyang Lin. 2024.
\newblock \href {https://arxiv.org/abs/2410.23933} {Language models can self-lengthen to generate long texts}.

\bibitem[{{Qwen Team}(2025)}]{qwen2025qwen25}
{Qwen Team}. 2025.
\newblock \href {https://arxiv.org/abs/2412.15115} {Qwen2.5 technical report}.

\bibitem[{Rasley et~al.(2020)Rasley, Rajbhandari, Ruwase, and He}]{rasley2020deepspeed}
Jeff Rasley, Samyam Rajbhandari, Olatunji Ruwase, and Yuxiong He. 2020.
\newblock Deepspeed: System optimizations enable training deep learning models with over 100 billion parameters.
\newblock In \emph{Proceedings of the 26th ACM SIGKDD International Conference on Knowledge Discovery \& Data Mining}, pages 3505--3506.

\bibitem[{Rozière et~al.(2024)Rozière, Gehring, Gloeckle, Sootla, Gat, Tan, Adi, Liu, Sauvestre, Remez, Rapin, Kozhevnikov, Evtimov, Bitton, Bhatt, Ferrer, Grattafiori, Xiong, Défossez, Copet, Azhar, Touvron, Martin, Usunier, Scialom, and Synnaeve}]{rozière2024codellama}
Baptiste Rozière, Jonas Gehring, Fabian Gloeckle, Sten Sootla, Itai Gat, Xiaoqing~Ellen Tan, Yossi Adi, Jingyu Liu, Romain Sauvestre, Tal Remez, Jérémy Rapin, Artyom Kozhevnikov, Ivan Evtimov, Joanna Bitton, Manish Bhatt, Cristian~Canton Ferrer, Aaron Grattafiori, Wenhan Xiong, Alexandre Défossez, Jade Copet, Faisal Azhar, Hugo Touvron, Louis Martin, Nicolas Usunier, Thomas Scialom, and Gabriel Synnaeve. 2024.
\newblock \href {https://arxiv.org/abs/2308.12950} {Code llama: Open foundation models for code}.

\bibitem[{Sennrich et~al.(2016)Sennrich, Haddow, and Birch}]{sennrich2016improving}
Rico Sennrich, Barry Haddow, and Alexandra Birch. 2016.
\newblock \href {https://aclanthology.org/P16-1009} {Improving neural machine translation models with monolingual data}.
\newblock In \emph{Proceedings of the 54th Annual Meeting of the Association for Computational Linguistics (Volume 1: Long Papers)}.

\bibitem[{Shaham et~al.(2023)Shaham, Ivgi, Efrat, Berant, and Levy}]{shaham2023zeroscrolls}
Uri Shaham, Maor Ivgi, Avia Efrat, Jonathan Berant, and Omer Levy. 2023.
\newblock \href {https://aclanthology.org/2023.findings-emnlp.536} {{Z}ero{SCROLLS}: A zero-shot benchmark for long text understanding}.
\newblock In \emph{Findings of the Association for Computational Linguistics: EMNLP 2023}.

\bibitem[{Su et~al.(2024)Su, Ahmed, Lu, Pan, Bo, and Liu}]{su2024roformer}
Jianlin Su, Murtadha Ahmed, Yu~Lu, Shengfeng Pan, Wen Bo, and Yunfeng Liu. 2024.
\newblock \href {https://www.sciencedirect.com/science/article/pii/S0925231223011864} {Roformer: Enhanced transformer with rotary position embedding}.
\newblock \emph{Neurocomputing}.

\bibitem[{Touvron et~al.(2023)}]{touvron2023llama2}
Hugo Touvron et~al. 2023.
\newblock \href {https://arxiv.org/abs/2307.09288} {Llama 2: Open foundation and fine-tuned chat models}.

\bibitem[{Wang et~al.(2023)Wang, Kordi, Mishra, Liu, Smith, Khashabi, and Hajishirzi}]{wang2023instruct}
Yizhong Wang, Yeganeh Kordi, Swaroop Mishra, Alisa Liu, Noah~A. Smith, Daniel Khashabi, and Hannaneh Hajishirzi. 2023.
\newblock \href {https://aclanthology.org/2023.acl-long.754} {Self-instruct: Aligning language models with self-generated instructions}.
\newblock In \emph{Proceedings of the 61st Annual Meeting of the Association for Computational Linguistics (Volume 1: Long Papers)}.

\bibitem[{Xiong et~al.(2024)Xiong, Liu, Molybog, Zhang, Bhargava, Hou, Martin, Rungta, Sankararaman, Oguz, Khabsa, Fang, Mehdad, Narang, Malik, Fan, Bhosale, Edunov, Lewis, Wang, and Ma}]{xiong2024effective}
Wenhan Xiong, Jingyu Liu, Igor Molybog, Hejia Zhang, Prajjwal Bhargava, Rui Hou, Louis Martin, Rashi Rungta, Karthik~Abinav Sankararaman, Barlas Oguz, Madian Khabsa, Han Fang, Yashar Mehdad, Sharan Narang, Kshitiz Malik, Angela Fan, Shruti Bhosale, Sergey Edunov, Mike Lewis, Sinong Wang, and Hao Ma. 2024.
\newblock \href {https://aclanthology.org/2024.naacl-long.260} {Effective long-context scaling of foundation models}.
\newblock In \emph{Proceedings of the 2024 Conference of the North American Chapter of the Association for Computational Linguistics: Human Language Technologies (Volume 1: Long Papers)}.

\end{thebibliography}

\appendix
\clearpage
\section{Data format for NIAH Test Variants}
\label{sec:niah-data-format}
In Table~\ref{tab:niah-template}, we present the data format for NIAH tasks.
For more details, please refer to the original paper~\cite{hsieh2024ruler}.

\begingroup
\renewcommand{\arraystretch}{1.3}
\begin{table*}[]
    \footnotesize
    \centering
    \resizebox{\linewidth}{!}{
    \begin{tabular}{cp{0.95\linewidth}}
    \toprule
    \begin{tabular}{@{}c@{}}Single NIAH\end{tabular} & 
    \begin{tabular}{@{}p{\linewidth}@{}} 
    \textbf{Context:} \\
    Some special magic numbers are hidden within the following text. Make sure to memorize it. I will quiz you about the numbers afterwards.\\
    \textcolor{lightgray}{Paul Graham Essays.} \\
    \textcolor{lightgray}{......} One of the special magic numbers for \textcolor{violet}{\{word\}} is: \textcolor{orange}{\{uuid\}}. \textcolor{lightgray}{......}\\
    What is the special magic number for \textcolor{violet}{\{word\}} mentioned in the provided text? \\ \\
    \textbf{Answer Prefix:} \\
    The special magic number for \textcolor{violet}{\{word\}} mentioned in the provided text is\end{tabular}\\

    \midrule

    \begin{tabular}{@{}c@{}}Multi-key NIAH\end{tabular} & 
    \begin{tabular}{@{}p{\linewidth}@{}} 
    \textbf{Context:} \\
    Some special magic uuids are hidden within the following text. Make sure to memorize it. I will quiz you about the uuids afterwards.\\
    \textcolor{lightgray}{Paul Graham Essays.} \\
    \textcolor{lightgray}{One of the special magic uuids for \{word-1\} is: \{uuid-1\}.} \\
    \textcolor{lightgray}{One of the special magic uuids for \{word-2\} is: \{uuid-2\}.} \\
    \textcolor{lightgray}{......} One of the special magic uuids for \textcolor{violet}{\{word-x\}} is: \textcolor{orange}{\{uuid-x\}}. \textcolor{lightgray}{......}\\
    \textcolor{lightgray}{One of the special magic uuids for \{word-n-1\} is: \{uuid-n-1\}.} \\
    \textcolor{lightgray}{One of the special magic uuids for \{word-n\} is: \{uuid-n\}.} \\
    What is the special magic number for \textcolor{violet}{\{uuid-x\}} mentioned in the provided text? \\ \\
    \textbf{Answer Prefix:} \\
    The special magic number for \textcolor{violet}{\{uuid-x\}} mentioned in the provided text is\end{tabular}\\

    \midrule
    
    \begin{tabular}{@{}c@{}}Multi-query NIAH\end{tabular} & 
    \begin{tabular}{@{}p{\linewidth}@{}} 
    \textbf{Context:} \\
    Some special magic numbers are hidden within the following text. Make sure to memorize it. I will quiz you about the numbers afterwards.\\
    \textcolor{lightgray}{Paul Graham Essays.} \\
    \textcolor{lightgray}{......} One of the special magic numbers for \textcolor{violet}{\{word-1\}} is: \textcolor{orange}{\{uuid-1\}}. \textcolor{lightgray}{......} \\    
    \textcolor{lightgray}{......} One of the special magic numbers for \textcolor{violet}{\{word-2\}} is: \textcolor{orange}{\{uuid-2\}}. \textcolor{lightgray}{......} \\
    \textcolor{lightgray}{......} One of the special magic numbers for \textcolor{violet}{\{word-3\}} is: \textcolor{orange}{\{uuid-3\}}. \textcolor{lightgray}{......} \\
    \textcolor{lightgray}{......} One of the special magic numbers for \textcolor{violet}{\{word-4\}} is: \textcolor{orange}{\{uuid-4\}}. \textcolor{lightgray}{......}\\
    What are all the special magic numbers for \textcolor{violet}{\{word-1\}}, \textcolor{violet}{\{word-2\}}, \textcolor{violet}{\{word-3\}}, and \textcolor{violet}{\{word-4\}} mentioned in the provided text? \\ \\
    \textbf{Answer Prefix:} \\
    The special magic numbers for \textcolor{violet}{\{word-1\}}, \textcolor{violet}{\{word-2\}}, \textcolor{violet}{\{word-3\}}, and \textcolor{violet}{\{word-4\}} mentioned in the provided text are\end{tabular}\\

    \midrule
    
    \begin{tabular}{@{}c@{}}Multi-value NIAH\end{tabular} & 
    \begin{tabular}{@{}p{\linewidth}@{}} 
    \textbf{Context:} \\
    Some special magic numbers are hidden within the following text. Make sure to memorize it. I will quiz you about the numbers afterwards.\\
    \textcolor{lightgray}{Paul Graham Essays.} \\
    \textcolor{lightgray}{......} One of the special magic numbers for \textcolor{violet}{\{word\}} is: \textcolor{orange}{\{uuid-1\}}. \textcolor{lightgray}{......}\\
    \textcolor{lightgray}{......} One of the special magic numbers for \textcolor{violet}{\{word\}} is: \textcolor{orange}{\{uuid-2\}}. \textcolor{lightgray}{......}\\
    \textcolor{lightgray}{......} One of the special magic numbers for \textcolor{violet}{\{word\}} is: \textcolor{orange}{\{uuid-3\}}. \textcolor{lightgray}{......}\\
    \textcolor{lightgray}{......} One of the special magic numbers for \textcolor{violet}{\{word\}} is: \textcolor{orange}{\{uuid-4\}}. \textcolor{lightgray}{......}\\
    What are all the special magic numbers for \textcolor{violet}{\{word\}} mentioned in the provided text? \\ \\
    \textbf{Answer Prefix:} \\
    The special magic numbers for \textcolor{violet}{\{word\}} mentioned in the provided text are\end{tabular}\\
    \bottomrule
    \end{tabular}}
    \caption{Data formats for different NIAH tasks: Single NIAH, Multi-key NIAH, Multi-query NIAH and Multi-value NIAH.}
    \label{tab:niah-template}
\end{table*}
\endgroup

\section{Template for Instruction Synthesis}
\begin{figure}[htb]
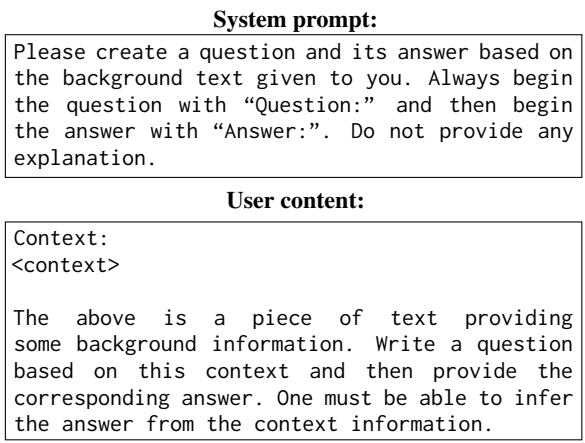

  \centering\small
  \textbf{System prompt:}
  \vspace{1ex}
  \noindent\framebox{%
  \parbox{0.46\textwidth}{
  \texttt{Please create a question and its answer based on the background text given to you. Always begin the question with ``Question:'' and then begin the answer with ``Answer:''. Do not provide any explanation.}
  }%
  }
  \vspace{1ex}
  \textbf{User content:}
  \vspace{1ex}
  \noindent\framebox{%
  \parbox{0.46\textwidth}{
  \texttt{Context:\\
  <context>\\
  \\
  The above is a piece of text providing some background information. Write a question based on this context and then provide the corresponding answer. One must be able to infer the answer from the context information.}
  }%
  }
\caption{The prompt template for synthesizing an instruction-answer pair from a given context. The template takes a context passage as input, where \texttt{<context>} is are replaced with the actual context text. The system prompt ensures the output follows a consistent format, while the user content guides the LLM to generate instruction-answer pair based on the given context.}
\label{fig:prompt-template-qa-synthesis}
\end{figure}

\begin{figure}[ht!]
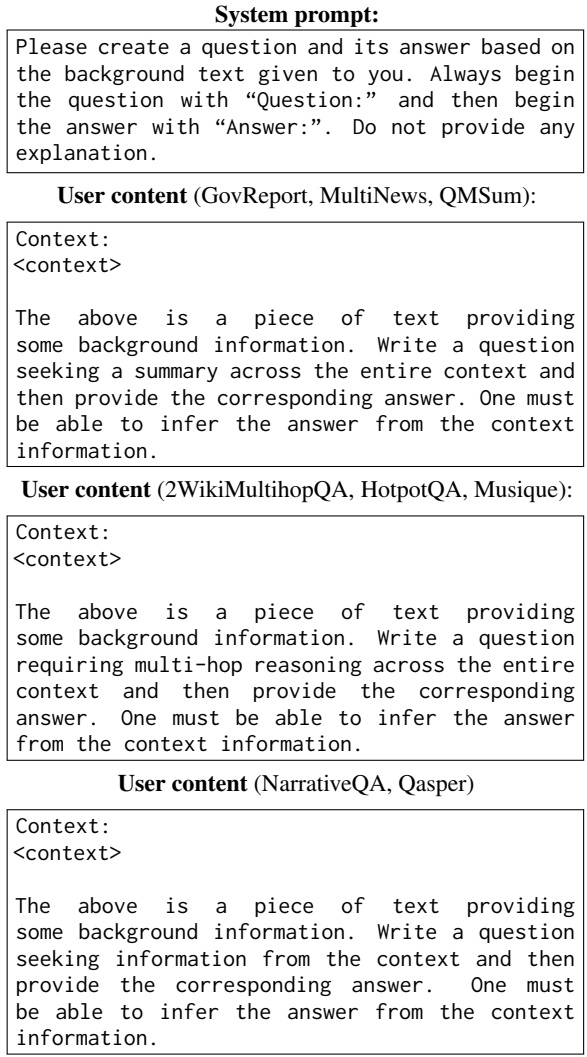

  \centering\small
  \textbf{System prompt:}
  \vspace{1ex}
  \noindent\framebox{%
  \parbox{0.46\textwidth}{
  \texttt{Please create a question and its answer based on the background text given to you. Always begin the question with ``Question:'' and then begin the answer with ``Answer:''. Do not provide any explanation.}
  }%
  }
  \vspace{1ex}
  \textbf{User content} (GovReport, MultiNews, QMSum):
  \vspace{1ex}
  \noindent\framebox{%
  \parbox{0.46\textwidth}{
  \texttt{Context:\\
  <context>\\
  \\
  The above is a piece of text providing some background information. Write a question seeking a summary across the entire context and then provide the corresponding answer. One must be able to infer the answer from the context information.}
  }%
  }
  \vspace{1ex}
  \textbf{User content} (2WikiMultihopQA, HotpotQA, Musique):
  \vspace{1ex}
  \noindent\framebox{%
  \parbox{0.46\textwidth}{
  \texttt{Context:\\
  <context>\\
  \\
  The above is a piece of text providing some background information. Write a question requiring multi-hop reasoning across the entire context and then provide the corresponding answer. One must be able to infer the answer from the context information.}
  }%
  }
  \vspace{1ex}
  \textbf{User content} (NarrativeQA, Qasper)
  \vspace{1ex}
  \noindent\framebox{%
  \parbox{0.46\textwidth}{
  \texttt{Context:\\
  <context>\\
  \\
  The above is a piece of text providing some background information. Write a question seeking information from the context and then provide the corresponding answer. One must be able to infer the answer from the context information.}
  }%
  }
\caption{The task-constrained prompt template for synthesizing an instruction-answer pair from a given context.}
\label{fig:prompt-task-template-context-synthesis}
\end{figure}

For instruction synthesis, we adopt the template shown in Figure~\ref{fig:prompt-template-qa-synthesis}, which prompts the model to generate instruction-answer pairs from given context.
While this template does not constrain the instruction type, we also experiment with a task-specific template (Figure~\ref{fig:prompt-task-template-context-synthesis}) that explicitly specifies the instruction type - generating summarization instructions for summarization tasks (GovReport, MultiNews, QMSum), generating multi-hop questions for multi-document QA tasks (2WikiMultihopQA, HotpotQA, Musique) and generating single-hop questions for single-document QA tasks (NarrativeQA, Qasper).
As shown in Table~\ref{tab:performance-task-specific-template}, the template (Figure~\ref{fig:prompt-template-qa-synthesis}) we apply in our main experiments produce higher performance.

\section{Examples of Synthesized Context}
\label{sec:example-synthesized-context}
We present examples of our synthesized context from question-answering tasks (Table~\ref{tab:synthesis-context-qa-case}) and summarization tasks (Table~\ref{tab:synthesis-context-sum-case}) to help readers better understand the benefit of our approach.
Taking the first example in Table~\ref{tab:synthesis-context-qa-case}, the evidence in the synthesized context is distributed across different parts of the text, while detailed background information serve as challenging distractors.
We suggest that this context composition helps the model learn robust patterns for context utilization.

\begingroup
\renewcommand{\arraystretch}{1.3}
\begin{table*}[]
    \scriptsize
    \centering
    \resizebox{\linewidth}{!}{
    \begin{tabular}{p{0.95\linewidth}}
    \toprule
    \textbf{<Synthesized Context>} \\
    Harry Kane, born on July 28, 1993, in Walthamstow, London, is a prominent English footballer known for his impressive goal-scoring ability and leadership on the pitch. From a young age, Kane displayed a passion for football and began his journey in the sport through local youth teams. He joined the Tottenham Hotspur academy at the age of 11, where he developed his skills and nurtured his ambition to become a professional footballer.\\
    Kane's rise through the ranks was not without challenges. He faced setbacks including being loaned out to various lower-league clubs like Leyton Orient, Millwall, Norwich City, and Leicester City, where he honed his skills but struggled to secure a permanent spot. Despite these difficulties, Kane's determination and work ethic paid off when he finally broke into the Tottenham first team during the 2014-2015 season. \\
    As a forward, Kane quickly established himself as a key player for Tottenham, displaying a remarkable ability to score goals from various positions on the pitch. His playing style is characterized by a combination of technical skill, tactical intelligence, and physicality. Additionally, he is known for his versatility, being able to play both as a striker and as a deeper-lying forward, contributing assists to his teammates. \\
    Kane's accomplishments at club level have earned him numerous accolades, including multiple Premier League Golden Boot awards, recognizing him as the league's top scorer. His remarkable performances have significantly contributed to Tottenham's competitiveness in both the Premier League and European competitions. \\
    In parallel with his club success, Kane has also made significant contributions to the England national team. He made his senior debut in 2015 and quickly became an integral part of the national squad. His leadership qualities shone through when he was appointed the captain of the national team, leading them through pivotal matches, including the FIFA World Cup 2018, where he finished as the tournament's top scorer. \\
    Throughout his career, Kane has been associated with Nike, opting to wear the Nike Hypervenom football boot, which is designed for agility and precision - a perfect match for his style of play. His affinity for these boots is a testament to his competitive edge and dedication to performing at his best. \\
    As Kane progresses in his career, he continually strives to achieve greater heights, including aspirations for club trophies and international honors. The expectations placed on him by fans, media, and himself create an environment filled with pressure, yet he remains focused on his ambitions. Balancing his professional responsibilities with personal life, Kane shares experiences with his family, which provides him with support and motivation. \\ 
    With a strong relationship with his coach and teammates, Kane fosters an atmosphere of camaraderie within the squad, encouraging younger players and leading by example. His professionalism and commitment to the sport have made him a role model for aspiring footballers across the globe. \\
    Through his journey, Kane embodies the spirit of perseverance, determination, and excellence, showcasing what it means to be a professional athlete in today's competitive sports landscape. \\
    \textbf{<Instruction>} \\
    English footballer, Harry Kane, wears the Nike Hypervenom football boot, and also plays as a forward for what teams? \\
    \textbf{<Answer>} \\
    Premier League club Tottenham Hotspur and the England national team \\
    \hline
    \textbf{<Synthesized Context>} \\
    In a recent study on the efficacy of automated question-answering systems, researchers were tasked with assessing the accuracy of a dataset compiled from various sources. The dataset consisted of 1,000 question-answer pairs related to multiple subjects, including science, history, technology, and culture. Initially, the dataset was created using a mix of crowd-sourced responses and automatically generated answers. To ensure the quality and reliability of the information, the research team set out to evaluate the correctness of each pair. \\
    To do this, they employed a team of six annotators - graduate students from several disciplines, including computer science, linguistics, and education. The annotators were trained to review the question-answer pairs based on a detailed rubric developed by the researchers. This rubric included criteria such as factual accuracy, relevance to the question, clarity of the answer, and overall coherence. \\
    The annotators were tasked with individually assessing each question-answer pair and providing feedback. After the individual evaluations, the team convened to discuss their findings and reach a consensus on the correctness of each pair. A crucial part of this process was the inter-annotator agreement, which measured the level of agreement among the annotators regarding the correctness of the pairs. \\
    Upon reviewing the pairs, it was found that the inter-annotator agreement was high, with an average agreement score of 85\%. This indicated a strong level of consistency among annotators. To further ensure quality, it was stipulated that the correctness of all the question-answer pairs had to be verified by at least two annotators before they could be deemed reliable. This measure was put in place to eliminate any potential biases and discrepancies that may arise from individual assessments. \\
    The study aimed to explore not only the reliability of the datasets used in existing automated systems but also the quality of answers provided by such systems. The goal was to contribute meaningful insights into the development of more effective AI tools that could provide accurate and relevant information to users. The researchers recognized the challenges posed by the rapidly changing nature of knowledge and how that impacts the creation of question-answer pairs. \\
    Ultimately, the reliability of the dataset was crucial, as it would be employed in further testing of the automated systems, aiming to improve their accuracy and user satisfaction. The research team hoped to publish their findings in a peer-reviewed journal, contributing to the field of artificial intelligence and educational technology by highlighting the importance of quality control in data used for training machine learning models. \\
    \textbf{<Instruction>} \\
    What was the inter-annotator agreement? \\
    \textbf{<Answer>} \\
    Correctness of all the question answer pairs are verified by at least two annotator. \\
    \bottomrule
    \end{tabular}
    }
    \caption{Examples of the synthesized background context for instruction-answer pairs generated by \texttt{GPT4o-mini} in the question-answering task. For the sake of space, we only show the relevant background context. In practice, additional unrelevant context is concatenated to reach longer length.}
    \label{tab:synthesis-context-qa-case}
\end{table*}
\endgroup

\begingroup
\renewcommand{\arraystretch}{1.3}
\begin{table*}[ht]
    \scriptsize
    \centering
    \resizebox{\linewidth}{!}{
    \begin{tabular}{p{0.95\linewidth}}
    \toprule
    \textbf{<Synthesized Context>} \\
    In the early 2010s, the American public was becoming increasingly aware of the growing obesity epidemic and its links to unhealthy eating habits. First Lady Michelle Obama, concerned about the health of children and families in the United States, initiated the \"Let's Move!\" campaign in 2010 aimed at reducing childhood obesity and promoting healthy eating and physical activity. The focus was on increasing access to healthy foods, particularly in low-income communities, where access to fresh produce was often limited due to food deserts.\\
    During this period, Wal-Mart, the largest retailer in the United States, was facing criticism for contributing to unhealthy eating habits through the low-cost, processed foods it sold. Recognizing its potential to influence consumer choices and health outcomes significantly, Wal-Mart sought to improve its image and address these health concerns. The company realized that by modifying its product offerings and focusing on healthier options, it could not only contribute to the public health initiative but also capture a growing market of health-conscious consumers.
    Wal-Mart's CEO and management team held strategic meetings to explore how to implement healthier food options across their stores. They acknowledged that despite Wal-Mart's low prices being praised by many, the foods that gained the most sales - often high in sodium, trans fats, and sugars - had detrimental health effects, particularly for families with limited budgets. They recognized their unique position to make a difference due to their extensive reach.
    Meanwhile, Michelle Obama was seeking partnerships with major corporations to extend the reach of her campaign. She viewed initiatives with retailers like Wal-Mart as instrumental in changing the food landscape in America. After several negotiations and discussions, it was announced that Wal-Mart would reformulate their private label, Great Value, to reduce unhealthy ingredients and increase the availability of fruits, vegetables, and whole grains at lower prices.
    This partnership was publicized at a high-profile press conference attended by media outlets, health food advocates, and community leaders, including Michelle Obama. During her speech, Obama emphasized the importance of better nutrition for children and applauded Wal-Mart's commitment to making healthier options more accessible. Her statement, ``It's a victory for parents. It's a victory for families, but most of all, it's a victory for our children,'' encapsulated the larger narrative that this initiative was more than just a business decision; it was a movement towards a healthier future.
    Reports from industry analysts indicated that Wal-Mart's decision could significantly shape the market, particularly among low-income consumers who relied on Wal-Mart for affordable groceries. With the announcement, there was hope that similar efforts would follow suit from other retailers, leading to widespread changes in food production and sales practices.
    In terms of relationships, this initiative marked a pivotal point in the collaboration between corporate America and government health initiatives, showcasing how partnerships could be leveraged for public good. The dialogue surrounding the announcement also highlighted the shift in tone from the Obama administration towards businesses, encouraging them to participate in efforts that align with public health goals.
    Overall, this movement was positioned as a crucial step in addressing the health crisis affecting a significant portion of the American population, particularly children, and it sparked broader discussions about food justice, corporate responsibility, and the role of major retailers in public health initiatives. \\
    \textbf{<Instruction>} \\
    What is a summary of the news article? \\
    \textbf{<Answer>} \\
    Wal-Mart is embarking on a plan that could have a big impact on American nutrition: It will modify its house brand to make the food healthier, and lower costs for items such as fruits and veggies, and products made with whole grain. The initiative is the result of collaboration with Michelle Obama, reports the New York Times. \"It's a victory for parents. It's a victory for families, but most of all, it's a victory for our children,\" she said at today's announcement, notes ABC News. Industry analysts say that because of its reach, Wal-Mart has almost as much power as federal regulators to shape the market. Crucially, Wal-Mart, and specifically its house brand Great Value, are popular with low-income shoppers - the demographic that eats the most foods high in sodium, trans fats, and refined sugars. Related stories: Washngton Post: The move is a sign of the Obama administration's new tone toward corporate America in general and Wal-Mart in particular. Huffington Post: This could genuinely save lives, writes Dr. Michael F. Jacobson. \\
    \bottomrule
    \end{tabular}
    }
    \caption{An example of the synthesized background context for instruction-answer pairs generated by \texttt{GPT4o-mini} in the summarization task. For the sake of space, we only show the relevant background context. In practice, additional unrelevant context is concatenated to reach longer length.}
    \label{tab:synthesis-context-sum-case}
\end{table*}
\endgroup

\begingroup
\renewcommand{\arraystretch}{1.2}
\begin{table*}[ht]
\centering
\scriptsize
\resizebox{\textwidth}{!}{
\begin{tabular}{lccccccccc}
\hline
\textbf{\hspace{1cm}Instruction Data} & \textbf{NarrativeQA} & \textbf{Qasper}      & \textbf{HotpotQA}    & \textbf{2WikiQA}    & \textbf{MuSiQue}     & \textbf{GovReport} & \textbf{QMSum}       & \textbf{MultiNews} & \textbf{Avg.}        \\
\hline
\texttt{\hspace{1cm}LLaMA3.1-8B}               &                 &            &           &            &               &                 &                &                  &                      \\
UltraChat~\cite{ding2023enhancing}             & 22.45                & 28.12           & 24.00          & 19.38           & 9.08               & 30.24                & 26.18                & 27.36                & 23.35                \\
+ Instruction Synthesis (template {\small\ding{172}})  & 24.39                & 29.32           & 30.26          & 21.68           & 14.99              & 29.85                & 25.60                & 27.02                & 25.39                \\
+ Instruction Synthesis (template {\small\ding{173}})  & 21.47                & 28.11           & 23.80          & 17.37           & 12.49              & 29.79                & 24.45                & 27.44                & 23.12                \\
+ Context Synthesis (ours)                     & \textbf{32.74}       & \textbf{45.30}  & \textbf{59.73} & \textbf{44.28}  & \textbf{32.20}     & \textbf{35.82}       & \textbf{27.79}       & \textbf{30.70}       & \textbf{38.57}       \\
\hline
\end{tabular}
}
\caption{This table compares model performance using different templates for instruction synthesis. Template {\small\ding{172}} refers to the constrain-free template in Figure~\ref{fig:prompt-template-qa-synthesis}, while Template {\small\ding{173}} refer to the task-specific template in Figure~\ref{fig:prompt-task-template-context-synthesis}.}
\label{tab:performance-task-specific-template}
\end{table*}
\endgroup

\section{Instruction-tuning Details}
\label{sec:instruction-tuning-details}
We use AdamW optimizer with $\beta_1$=0.9, $\beta_2$=0.95 for instruction-tuning. 
The learning rate is set to 2e-5 with a cosine decay schedule and 3\% warm-up ratio.
The models are fine-tuned for 2 epochs.
For training efficiency, we employ DeepSpeed ZeRO-3~\cite{rasley2020deepspeed} alongside a packing strategy with loss weighting~\cite{bai2024longalign}.
Specially, we adopt a packing strategy where training samples are packed together, with loss computed only on the output tokens. 
Following the default settings in \textit{LongAlign} codebase, we set the maximum sequence length per batched sample to 65536 and 32768 for \texttt{LLaMA2-7B-64k} and \texttt{LLaMA3.1-8B-128k} respectively.
Training is conducted on 8$\times$H800 GPUs with a per-device batch size of 1.

\section{Experimental Results with ShareGPT}
\label{sec:llama3.1-sharegpt}
Table~\ref{tab:sharegpt} presents the experimental results with ShareGPT on \texttt{LLaMA3.1-8B}. 
The results demonstrates similar findings to those observed in Table~\ref{tab:compare}.

\section{Impact of Context Concatenation Size}
\label{sec:number-of-concatenated-context}
In this paper, we set the concatenation size to ten, consisting of one relevant background context and nine irrelevant contexts.
Table~\ref{tab:number-of-concatenated-context} presents the performance results across different numbers of concatenated contexts.
Our results indicate that larger concatenation sizes generally yield better performance.

\begingroup
\renewcommand{\arraystretch}{1.2}
\begin{table*}[ht]
\centering
\scriptsize
\resizebox{\textwidth}{!}{
\begin{tabular}{lccccccccc}
\hline
\textbf{\hspace{0.6cm}Instruction Data}    & \textbf{NarrativeQA} & \textbf{Qasper}      & \textbf{HotpotQA}    & \textbf{2WikiQA}    & \textbf{MuSiQue}     & \textbf{GovReport} & \textbf{QMSum}       & \textbf{MultiNews} & \textbf{Avg.}        \\
\hline
\texttt{\hspace{0.6cm}LLaMA3.1-8B}         &                 &            &           &            &               &                 &                &                  &                      \\
ShareGPT~\cite{vicuna2023}               & 23.35           & 23.48      & 30.98     & 24.67      & 10.19         & 29.89           & 23.60          & 28.09            & 24.28                \\
+ Instruction Synthesis                  & 26.32           & 32.19             & 28.80          & 28.86           & 14.81             & 30.78            & 24.46           &  27.58         & 26.73                \\
+ Context Synthesis (ours)               & \textbf{31.16}  & \textbf{41.02}    & \textbf{54.27} & \textbf{38.92}  & \textbf{28.03}    & \textbf{35.06}   & \textbf{26.99}  & \textbf{31.02} & \textbf{35.81}       \\
\hline
\end{tabular}
}
\caption{This table illustrates model performance between using general instruction data (ShareGPT) alone and using additional long-context intruction data (rows with `+').}
\label{tab:sharegpt}
\end{table*}
\endgroup

\begingroup
\renewcommand{\arraystretch}{1.2}
\begin{table*}[ht]
\centering
\scriptsize
\resizebox{\textwidth}{!}{
\begin{tabular}{lccccccccc}
\hline
\textbf{\hspace{0.4cm}Instruction Data} & \textbf{NarrativeQA} & \textbf{Qasper}      & \textbf{HotpotQA}    & \textbf{2WikiQA}    & \textbf{MuSiQue}     & \textbf{GovReport} & \textbf{QMSum}       & \textbf{MultiNews} & \textbf{Avg.}        \\
\hline
\texttt{\hspace{0.6cm}LLaMA3.1-8B}          &                 &                &                &                 &               &                 &                &                 &                 \\
UltraChat~\cite{ding2023enhancing}             & 22.45           & 28.12          & 24.00          & 19.38           & 9.08          & 30.24           & 26.18          & 27.36           & 23.35           \\
+ Context Synthesis (n=1)               & 32.61           & 44.46          & 58.08          & 38.35           & \textbf{32.91} & 35.81          & 27.10          & \textbf{32.03}  & 37.67           \\
+ Context Synthesis (n=5)               & 32.10           & 42.08          & \textbf{60.15} & \textbf{45.34}  & 30.15          & \textbf{35.98} & 26.51          & 31.04           & 37.92           \\
+ Context Synthesis (n=10)              & \textbf{32.74}  & \textbf{45.30} & 59.73          & 44.28           & 32.20          & 35.82          & \textbf{27.79} & 30.70           & \textbf{38.57}  \\
\hline
\end{tabular}
}
\caption{Performance comparison across different concatenation sizes (n). n represents the number of concatenated contexts, including one relevant context and n-1 irrelevant contexts. Results show that larger concatenation sizes generally lead to better overall performance.}
\label{tab:number-of-concatenated-context}
\end{table*}
\endgroup

\begin{figure*}[ht!]
    \centering
    \includegraphics[width=\linewidth]{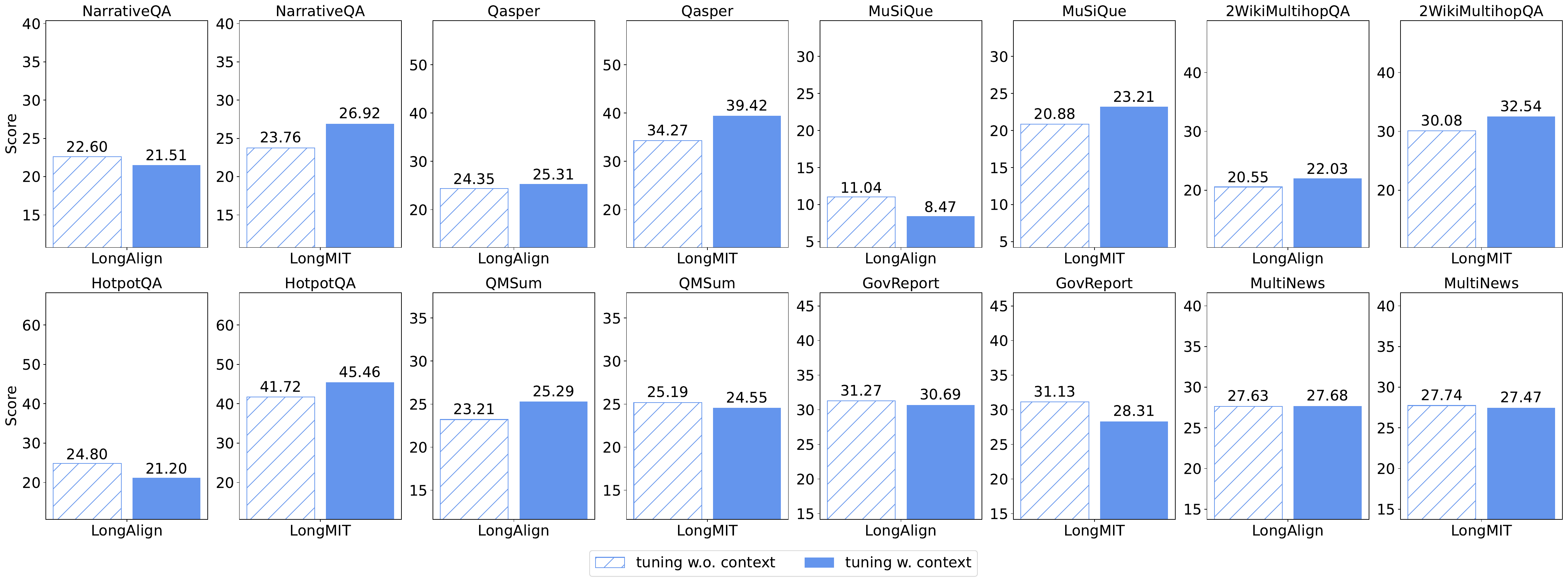}
    \caption{In this figure we compare tuning without context (diagonal lines) with tuning with context (solid bars) and assess context-instruction coherence in synthetic instruction data (LongAlign, LongMIT). Experiments are conducted with \texttt{LLaMA3.1-8B}.}
    \label{fig:longalign-longmit}
\end{figure*}

\section{Context-Instruction Coherence Analysis}
\label{sec:longalign-longmit-context-free-tuning}
With our proposed analytic tool, we measure the context-instruction coherence in synthetic instruction data (LongAlign, LongMIT) by previous instruction sysnthesis approaches.
Experimental results are depicted in Figure~\ref{fig:longalign-longmit}.
For LongAlign, we observe minimal difference between context-included and context-free tuning, suggesting poor context-instruction coherence in their synthetic data.
While LongMIT enhances the quality of synthetic data through a carefully designed multi-agent workflow for question-answering tasks, it has limited generalizability across different tasks and achieves lower performance compared to our approach.

\section{Used Scientific Artifacts}
Below lists scientific artifacts that are used in our work. For the sake of ethic, our use of these artifacts is consistent with their intended use.
\begin{itemize} [itemsep=1pt]
    \item \textit{LongAlign (Apache-2.0 license)}, a codebase developed for long-context instruction-tuning. 
    \item \textit{RULER (Apache-2.0 license)}, a repository for generating synthetic examples to evaluate long-context language models with configurable sequence length and task complexity. 
    \item \textit{LongBench (MIT license)}, a benchmark designed for assessing the long-context capabilities of large language models.
    \item \textit{ZeroScrolls (MIT license)}, a benchmark for evaluating the long-context capabilities of large language models.
    \item \textit{LLaMA2-7B-64K (Apache-2.0 license)}, a continued pretrained version of \texttt{LLaMA2-7B} with an extended 64k context window.
    \item \textit{LLaMA-3.1 (LLaMA3.1 license)}, a large language model developed by Meta. 
    \item \textit{GPT4o-mini (Proprietary license)}, a large language model developed by OpenAI.
    \item \textit{Qwen-2.5-72B (Qwen license)}, a large language model developed by Qwen.
\end{itemize}

\end{document}